\documentclass[letterpaper, 10 pt, conference]{ieeeconf}
\IEEEoverridecommandlockouts  
\usepackage{graphics} 
\usepackage{epsfig} 
\usepackage{amsmath} 
\usepackage{amssymb}  
\usepackage{booktabs}
\usepackage{xcolor}

\title{AirDreamer: Generalist Drone Navigation with World Models}

\author{Zian Liu$^{*,1}$, Andong Yang$^{*,2}$, Chunkai Yang$^{3}$, Ruidong An$^{1}$, Chao Gao$^{1}$, Guyue Zhou$^{1}$
\thanks{$^{*}$Equal contribution.}
\thanks{$^{1}$Institute for AI Industry Research, Tsinghua University, Beijing, China.}
\thanks{$^{2}$andongyang@mail.tsinghua.edu.cn, Department of Electronic Engineering, Tsinghua University, Beijing, China.}
\thanks{$^{3}$School of Remote Sensing and Information Engineering, Wuhan University, Wuhan, China.}
\thanks{Corresponding author: Chao Gao, Guyue Zhou.}
}

\begin{document}
\maketitle
\thispagestyle{empty}
\pagestyle{empty}

\begin{abstract}
Navigating a drone in unseen and cluttered environments requires reliable generalization to unseen scene layouts and understanding of environmental structure relative to the robot’s capabilities. Previous methods, which assume the same environment configuration, often rely heavily on human-designed perception pipelines and predefined rules to guide the robot toward the target. This process is environment-dependent and generalizes poorly across environments. Inspired by animal navigation behavior, we design a navigation framework that navigates with a reinforcement-learning-based policy on top of a world-model-based environment understanding to overcome these issues. In addition, a sparse reward function without hand-crafted shaping terms is designed to avoid local minima traps and encourage yaw control behaviors. In simulation and on real drones, our method exhibits emergent capabilities for navigating complex, unseen environments and escaping local optima where other methods fail. In challenging maps, it achieves a 5.3\%  higher navigation success rate than best baseline. Furthermore, the proposed framework achieves effective sim-to-real transfer without any tuning during deployment. The code will be publicly available.
\end{abstract}

\section{INTRODUCTION}
The capabilities of aerial robots have grown dramatically in recent years, showing potential in rescue, exploration, and transportation \cite{tranzatto2022cerberus}. However, deploying drones in unknown, cluttered environments remains challenging due to the wide variety of real-world conditions. To fully exploit the potential of aerial robots, developing emergent navigation behaviors in unknown, cluttered environments is a promising direction.

In the classical framework, environment and task understanding relies on the combination of perception and planning modules, which heavily depend on predefined heuristic functions or global information \cite{tordesillas2021faster}. A semantic or occupancy map generated by the perception module provides a searchable local environment representation \cite{zhou2020ego}. A hand-crafted cost function then guides the robot toward its destination. This cost function encodes the designer's understanding of the environment and task, but its form and weights are tuned empirically, with no guarantee that the resulting trajectory reflects the true navigation objective.

Learning-based navigation methods have emerged as a promising alternative. Existing learning-based navigation methods typically learn a tight coupling between sensor observations and actions  \cite{loquercio2021learning}. These methods are trained in simulation and suffer from out-of-distribution (OOD) failures, require costly sim-to-real tuning, and generalize poorly to unseen scenarios. Moreover, many learning-based approaches exhibit local-minimum behavior in non-convex environments, such as scenarios with large obstacles or traps, preventing successful navigation \cite{lee2025quadrotor,yu2024mavrllearnflycluttered}.

\begin{figure}[!t]
    \centering
    \vspace{6pt} 
    \includegraphics[width=8.1cm]{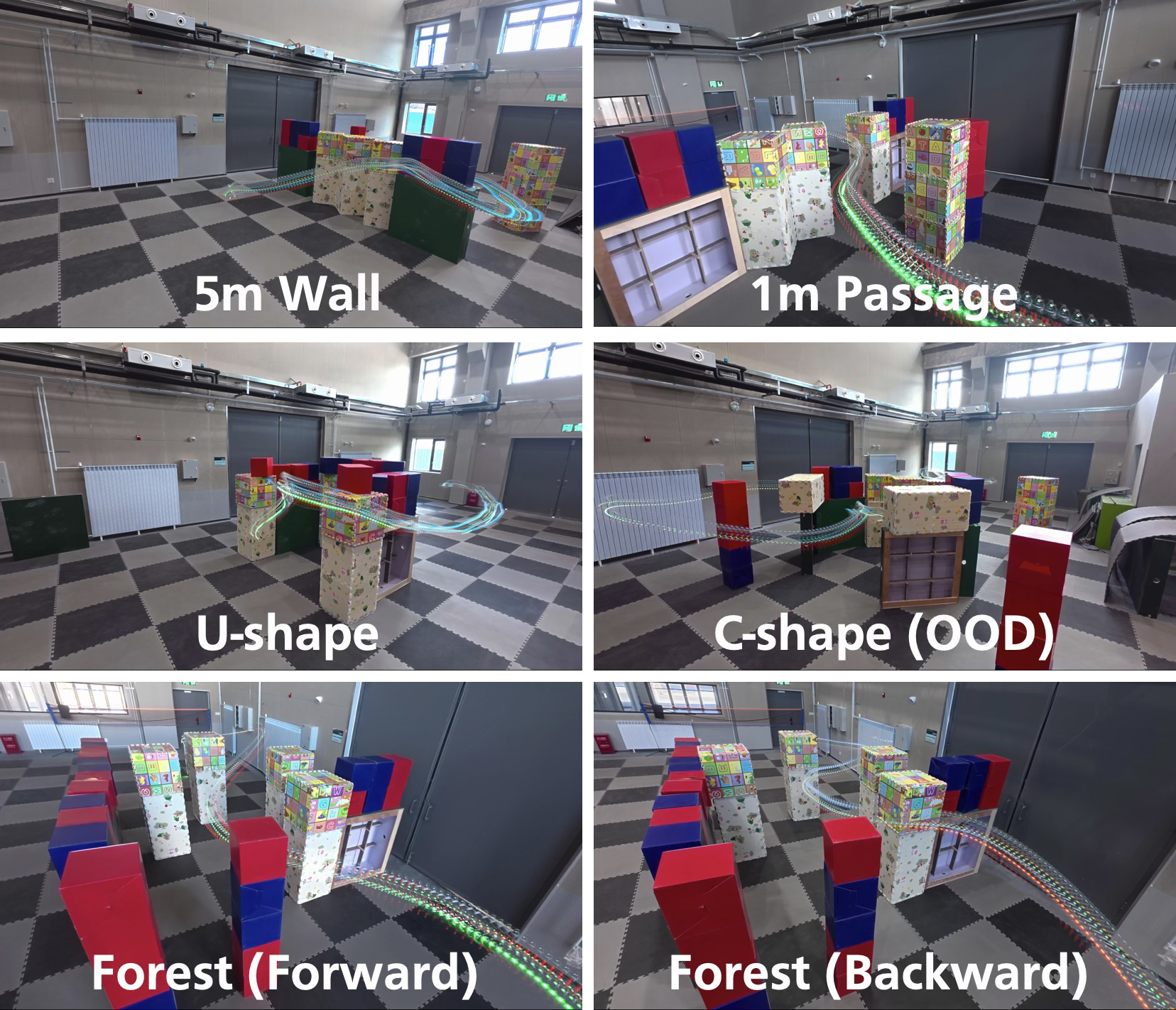}
    \caption{Autonomous navigation with the proposed AirDreamer in representative scenarios. On a real drone, AirDreamer successfully completes six challenging maps at speed up to 1.8 m/s and escapes from an out-of-distribution C-shaped obstacle.}
    \label{fig:cover}
\end{figure}

We take inspiration from rats that navigate around large and complex obstacles in unseen settings using local sensing and location. One explanation is that rats extract general patterns from their experiences and imagine rollouts of visited places to plan and improve their strategies \cite{pfeiffer2013hippocampal}. Therefore, rather than directly mapping sensor observations to actions, we build an internal understanding of the environment from raw observations and then make navigation decisions on top of this understanding. By decoupling environment understanding from action selection, our method achieves stronger robustness to OOD scenarios, smoother sim-to-real transfer, and better generalization across diverse environments.
    
Building on this principle, we propose AirDreamer, a vision-based drone navigation framework with two complementary components: a world model that encodes the environment from raw depth observations, and a reinforcement learning policy that reasons over this representation to select navigation actions. Since the world model captures environmental structure and provides dense predictions, the policy can be trained with a sparse reward, dramatically reducing reliance on hand-crafted guidance terms used in prior learning-based navigation methods \cite{yu2024mavrllearnflycluttered,lee2025quadrotor}. This design enables emergent navigation behaviors that escape local minima and generalize robustly across complex unseen environments. The main contributions of this work are:
    \begin{enumerate}
        \item A novel end-to-end aerial navigation framework is proposed that operates on onboard depth images and goal location alone, without pre-built maps or global obstacle information.
        \item A sparse reward is designed to reduce dense directional and hand-crafted shaping rewards to avoid reward hacking, remove human-imposed navigation preferences, and encourage yaw control behaviors that improve navigation performance. Such sparse rewards are typically infeasible for prior learning-based methods.   
        \item Extensive simulation and real-world experiments show that the proposed method navigates successfully in scenarios where prior methods fail, achieves state-of-the-art performance with a 5.3\% higher success rate than the best baseline, while requiring no tuning for deployment on real drones. The full system will be made publicly available to support further research in vision-based drone navigation.
    \end{enumerate} 

\section{RELATED WORK}
\subsection{Mobile Robot Navigation} 
Classical navigation systems typically adopt a cascaded scheme that partitions perception, mapping, planning, and control \cite{tordesillas2021faster}. Typically, sensor data are converted to point clouds and, together with pose estimates, aggregated into representations like grid maps or Euclidean signed distance fields (ESDFs) \cite{zhou2020ego}. A collision-free trajectory is then optimized with intermediate representations and tracked by a closed-loop controller \cite{kohler2025mpcframeworkefficientnavigation}. While interpretable and broadly effective, this pipeline is environment-dependent and compute-intensive, limiting its application in large environments. Additionally, its performance depends on each module's accuracy, hindering generalization beyond its designed setting.

Learning-based methods directly map observations to actions and avoid the error accumulation of modular pipelines. NavRL encodes lidar and depth into occupancy maps and obstacle states for a PPO policy \cite{xu2025navrllearningsafeflight}, but lacks historical context and struggles with large or non-convex obstacles. DepthNav targets such traps using training guidance \cite{lee2025quadrotor}. The policy is trained in a differentiable simulator with target-velocity guidance derived from a time-of-arrival gradient field, which is discarded at deployment so the policy internalizes the strategy. 
MAVRL \cite{yu2024mavrllearnflycluttered} targets cluttered environments by training an LSTM to predict future latent states within a VAE-encoded representation, equipping the model with memory and prediction. Shanks \textit{et al.} build a world-model-based navigation agent, but rely on a global map and an A* planner in a known environment and provide no real-world validation \cite{shanks2025dreamernav}. 

\subsection{World Models}
World models have emerged as a powerful paradigm for environment modeling in learning-based systems. Prediction-oriented world models focus on predicting action-conditioned future observations, as in Recurrent World Models \cite{ha2018world}, UniSim \cite{yang2024unisim}, and Genie \cite{bruce2024genie}. These models are primarily evaluated by prediction fidelity rather than downstream task performance.

World models have also been applied to embodied navigation and robot learning. Pathdreamer \cite{koh2021pathdreamer} and Navigation World Models \cite{bar2025navigation} predict future visual observations and use them to evaluate candidate actions. DayDreamer \cite{wu2023daydreamer} shows that Dreamer-style latent imagination works for online learning on real robots, and SkyDreamer \cite{verraest2025skydreamer} extends this approach to drone racing. However, these methods either rely on explicit planning over predicted observations or target structured settings such as race tracks, leaving end-to-end navigation in unknown, cluttered environments largely unexplored.

\section{PROBLEM FORMULATION}
This section formalizes vision-based goal navigation for a drone in unknown, cluttered environments with onboard depth sensing and computation. We define the coordinate frames, state the navigation objective, and model the task as a partially observable Markov decision process (POMDP).

\subsection{Coordinate Systems}
\begin{figure}[t]
    \centering
    \includegraphics[width=0.7\linewidth]{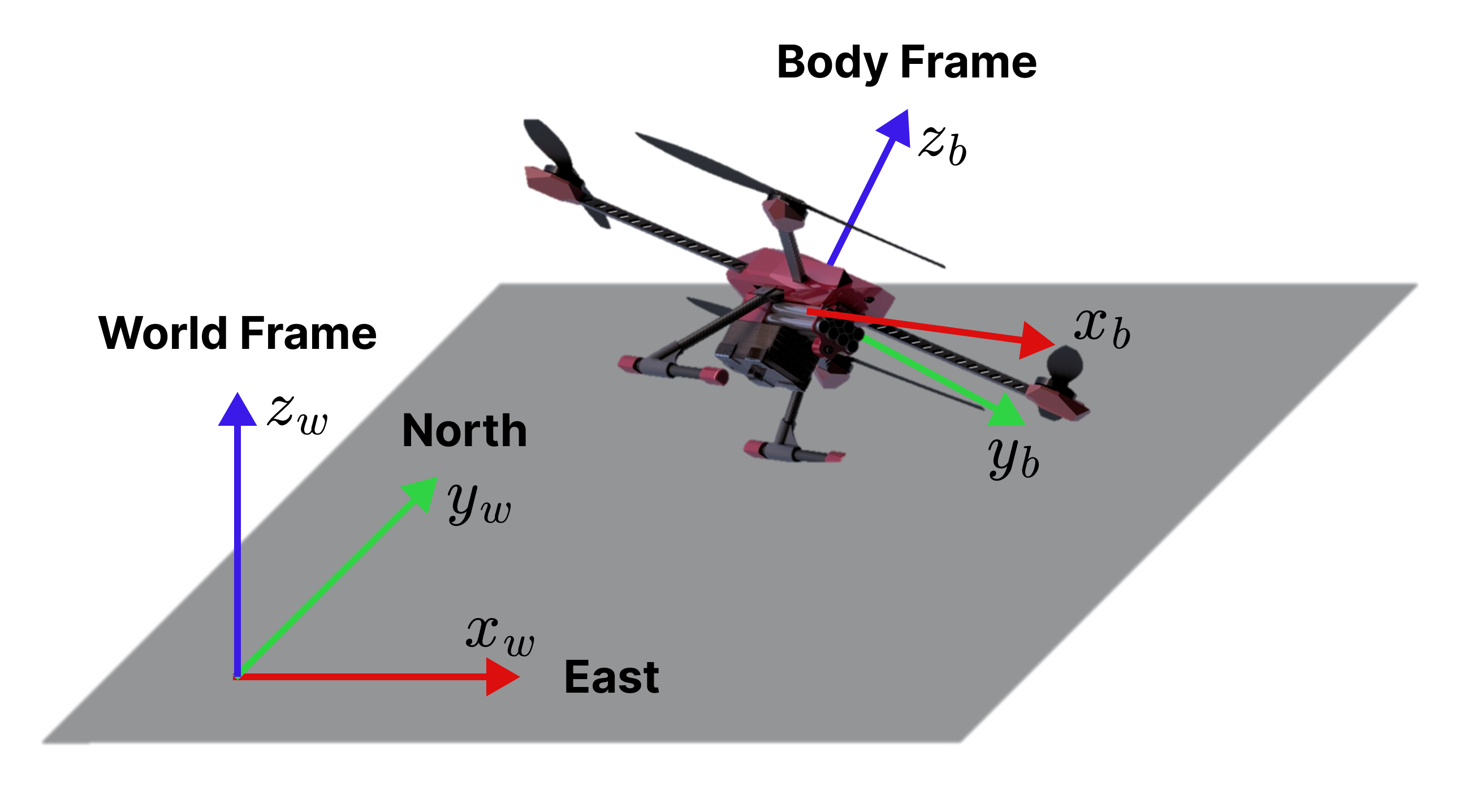}
    \caption{Reference frames used for AirDreamer. The body frame is attached to the drone, and the world frame is attached to the ground.}
    \label{fig:ref_sys}
\end{figure}
Two coordinate frames are used in this work. The first frame is the world reference frame $\mathbf{w}$ (ENU). It is an inertial frame fixed at a stationary point on the ground, in which absolute positions are defined and rewards are computed. The second frame is the body frame $\mathbf{b}$, with x-axis pointing towards the drone's nose and y-axis pointing to the left of the drone chassis. It is non-inertial but is convenient for defining camera-related quantities and local navigation states. We observe faster and more stable convergence during training when observations are expressed in this frame. A possible explanation is that doing so encourages the policy to learn direction-agnostic behaviors and reduces overfitting to the absolute layout of the simulation environment.

\subsection{Problem Statement}
\begin{figure}[t]
    \centering
    \includegraphics[width=0.86\linewidth]{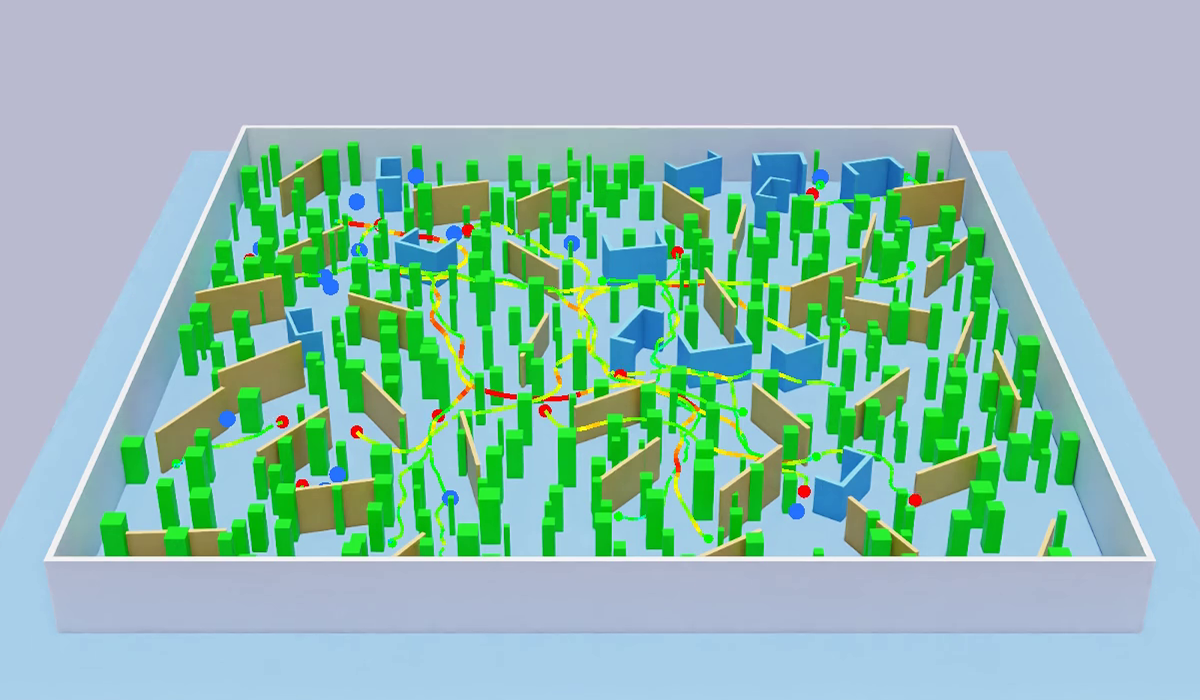}
    \caption{Example training environment.}
    \label{fig:env}
\end{figure}

The navigation task is defined as follows: the drone spawns at a random position and yaw, and must reach a random goal while maintaining altitude and avoiding obstacles. The observation consists of local information and location. We adopt a quadrotor as the experimental platform. Its 6-DoF dynamics and real-time control requirements provide a testbed for AirDreamer. We target settings where both traditional and previous learning-based methods fall into local optima or produce unsafe trajectories, such as non-convex obstacles, large obstacles, narrow gaps, clutter, and unseen layouts. These scenarios require long-horizon planning, non-greedy decision-making, and transferability of learned prior knowledge. Fig.~\ref{fig:env} shows one of the training environments in Isaac Sim. Random cuboid obstacles and walls teach general strategies for narrow gaps, clutter, and large convex obstacles, while U-shaped obstacles train the agent to escape from local traps and diversify priors for transfer.

We model navigation as a finite-horizon, partially observable MDP $(S, A, O, P, R, h, \gamma)$ over episode horizon $T$, with state space $S$, action space $A$, observation space $O$, transition $P(\mathbf{s}_{t+1} \mid \mathbf{s}_t, \mathbf{a}_t)$, reward $R$, sensor map $h(\mathbf{o}_t \mid \mathbf{s}_t)$, and discount $\gamma \in [0,1)$. The forward-facing depth camera leaves regions behind and to the sides unobserved, so the policy must infer them from history. A state
$\mathbf{s}_t = (\mathbf{p}_t, \mathbf{v}_t, \mathbf{q}_t, \boldsymbol{\Omega}_t, \mathbf{p}_g, \mathcal{M}, \mathbf{d})$
comprises world-frame position ($\mathrm{m}$) and linear velocity ($\mathrm{m/s}$), unit body-to-world quaternion, body-frame angular velocity ($\mathrm{rad/s}$), goal position, obstacle geometry and pose $\mathcal{M}$, and randomized drone parameters $\mathbf{d}$.

The action $\mathbf{a}_t = (\tilde{\mathbf{v}}^b_t, c_\Psi, s_\Psi) \in [-1,1]^5$ stacks a normalized body-frame velocity setpoint and a two-component yaw encoding that avoids radian discontinuities. Commands are decoded as
\begin{align}
    \mathbf{v}^{\,\mathrm{cmd}}_b &= \boldsymbol{\Lambda}\,\tilde{\mathbf{v}}^b_t,
        \quad \boldsymbol{\Lambda} = \mathrm{diag}(1.5, 1.5, 0.5)\,\mathrm{m/s}, \\
    \Psi^{\mathrm{cmd}} &= \mathrm{atan2}(s_\Psi, c_\Psi).
\end{align}
A low-level velocity controller tracks $(\mathbf{v}^{\,\mathrm{cmd}}_b, \Psi^{\mathrm{cmd}})$ while the policy issues commands at $20\,\mathrm{Hz}$.

The observation $(\mathbf{o}^{s}_t, \mathbf{o}^{d}_t) \sim h(\mathbf{s}_t)$ pairs a 15-dimensional state vector $\mathbf{o}^{s}_t$ with a depth image $\mathbf{o}^{d}_t \in [0.6, 6.0]^{48 \times 80}\,\mathrm{m}$:
\begin{equation}
    \mathbf{o}^{s}_t = [\hat{\mathbf{n}}^b_g, d_{xy}, d_z, \mathbf{v}^b_t, \boldsymbol{\Omega}_t, \mathbf{q}_t]^{\mathrm{T}},
\end{equation}
where $\hat{\mathbf{n}}^b_g$ is the unit goal direction in the body frame, and $d_{xy}, d_z$ are the horizontal goal distance and vertical offset. Neither $\mathcal{M}$ nor $\mathbf{d}$ is observed.

We seek a policy $\pi_\theta(\mathbf{a}_t \mid \mathbf{o}_0, \mathbf{a}_0, \dots, \mathbf{o}_t)$ conditioned on the history and maximizing the expected discounted return:
\begin{equation}
\label{eq:objective}
    \pi^\star_\theta = \arg\max_{\pi_\theta}\;
    \mathbb{E}\!\left[\sum_{t=0}^{T} \gamma^{\,t}\, R(\mathbf{s}_t, \mathbf{a}_t)\right].
\end{equation}

\section{METHODOLOGY}

\begin{figure*}[t]
    \centering
    \includegraphics[width=17.3cm]{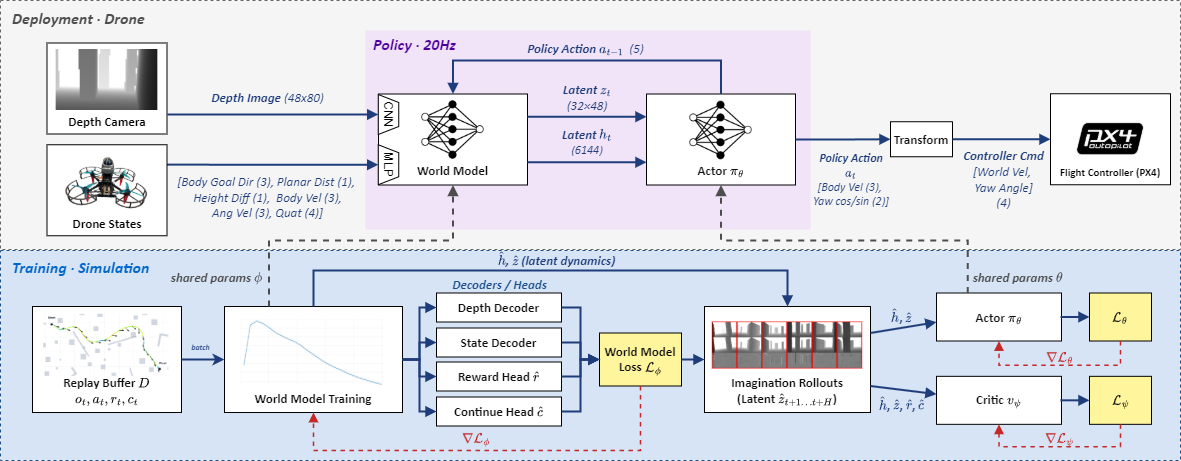}
    \caption{Architecture of AirDreamer. Modules used for real-world deployment are shown on a grey background. Policy training process in the simulator is shown on a blue background, comprising the replay buffer, decoders, and losses. Tensor dimensions are given in parentheses. Blue arrows denote data flow, red arrows denote gradient flow, and grey arrows denote model transfer from simulation to the real world.}
    \label{fig:arch}
\end{figure*}

Fig.~\ref{fig:arch} illustrates the architecture of AirDreamer. This work aims to enable drone navigation in complex environments using only an onboard depth camera and standard state estimation, without pre-built maps or hand-crafted heuristic information. The framework is trained end-to-end but architecturally decouples environment understanding from action generation, and a sparse reward is used for training. The flight controller, simulation environment, and real-world environment are detailed in Section \ref{sec:exp}.

\subsection{World Model and Policy}
The world model in AirDreamer understands the dynamics and the underlying environment principles relevant to the reward and task definitions. It uses the Dreamer V3 framework \cite{hafner2025mastering}. It encodes multimodal observations into compact latent states and predicts the next latent states, the reward, and the continuation probability conditioned on its memory and the current action. This is realized through an action-conditioned Recurrent State-Space Model (RSSM), defined in~\eqref{eq:rssm}. Trajectories collected through interaction are stored in a replay buffer $D$ for off-policy training.

\begin{equation} \label{eq:rssm}
\begin{aligned}
    \textit{Sequence Model:}     &\quad h_t = f_\phi(h_{t-1},\, z_{t-1},\, a_{t-1}) \\[4pt]
    \textit{Posterior:}          &\quad z_t \sim q_\phi(z_t \mid h_t,\, o_t) \\[4pt]
    \textit{Prior:}              &\quad \hat{z}_t \sim p_\phi(\hat{z}_t \mid h_t) \\[4pt]
    \textit{Reward Predictor:}   &\quad \hat{r}_t \sim p_\phi(\hat{r}_t \mid h_t,\, z_t) \\[4pt]
    \textit{Continue Predictor:} &\quad \hat{c}_t \sim p_\phi(\hat{c}_t \mid h_t,\, z_t) \\[4pt]
    \textit{Decoder:}            &\quad \hat{o}_t \sim p_\phi(\hat{o}_t \mid h_t,\, z_t)
\end{aligned}
\end{equation}

The RSSM's multimodal encoder receives both high-dimensional image inputs and low-dimensional sensor or state inputs: a convolutional encoder processes the normalized depth image $\mathbf{o}^d_t$, and a dense layer processes the symlog-transformed state vector $\mathbf{o}^s_t$; the two embeddings are concatenated before entering the RSSM. The model maps the POMDP inputs to Markovian latents through an RNN. It features a deterministic state $h_t \in \mathbb{R}^{6144}$ updated by a GRU and a stochastic state $z_t \in \{0,1\}^{32 \times 48}$, comprising $32$ categorical variables of $48$ classes each. The deterministic state is retained and updated by the RNN, which receives $a_{t-1}$ from the policy, and the stochastic state is produced from the deterministic state, and when real observations are available, also from the inputs. The decoder reconstructs the inputs from $z_t, h_t$; the reward and continuation heads feature twohot symexp categorical and Bernoulli distributions respectively. The encoder and decoder together form a variational encoder-decoder. In real-world execution, live observations are available, so the posterior is used.

The losses for training the world model are defined below. The prediction
loss $\mathcal{L}_\mathrm{pred}$ is the negative log-likelihood of the
observation, reward, and continuation under the posterior latent. The
dynamics loss $\mathcal{L}_\mathrm{dyn}$ trains the prior to match the
posterior, while the representation loss $\mathcal{L}_\mathrm{rep}$
regularises the posterior toward the prior. $\phi$ represents model weights.

\begin{equation} \label{eq:wm_loss}
\begin{aligned}
\mathcal{L}(\phi) = \mathbb{E}_{q_\phi}\!\Biggl[\sum_{t}\Bigl(
\mathcal{L}_{\mathrm{pred}}(\phi)
  + \mathcal{L}_{\mathrm{dyn}}(\phi) + 0.1\mathcal{L}_{\mathrm{rep}}(\phi)
\Bigr)\Biggr]
\end{aligned}
\end{equation}

\begin{figure*}[t]
    \centering
    \includegraphics[width=\linewidth]{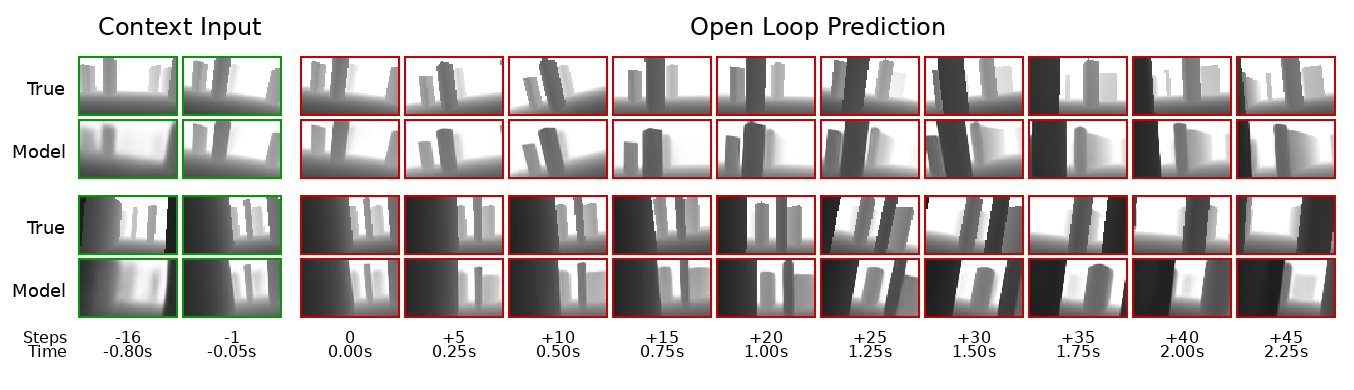}
    \caption{Latent imaginations of AirDreamer in the simulation. The images are decoded from the latent states $(h_t,\, z_t)$. Each rollout initializes $h_t$ with the context input and predicts future latent states without new observation under the current policy. The prediction produces sharp decoded images and maintains high visual fidelity for 20 steps, capturing both drone dynamics and environment layout.}
    \label{fig:dreaming}
\end{figure*}

The policy agent operates on the latent representation produced by the world model to select navigation actions, completing the decoupled framework. It adopts an actor-critic reinforcement learning structure. $\theta$ and $\psi$ are the weights of the actor and the critic.
\begin{equation} \label{eq:actor_critic}
\begin{aligned}
     \quad a_t \sim \pi_\theta(a_t \mid z_t, h_t) \quad \mu_t = \mathbb{E} \left[\pi_\theta \right]
\end{aligned}
\end{equation}

As shown in \eqref{eq:actor_critic}, the actor's policy $\pi_\theta$ takes the latent state $z_t$ and the recurrent hidden state $h_t$ from the world model as inputs and produces the action $a_t$, while the critic's value function $v_\psi(G_t \mid z_t, h_t)$ uses the same inputs to estimate the expected return $G_t$ via $\lambda$-return \eqref{eq:critic}.
\begin{equation} \label{eq:critic}
\begin{gathered}
    G_t^\lambda       = r_t + \gamma c_t \big((1 - \lambda)\, v_t + \lambda\, G_{t+1}^\lambda\big) \\
    G_T^\lambda       = v_T
\end{gathered}
\end{equation}

The actor serves as the policy agent and is trained exclusively on imagined trajectories generated using the prior, after the RSSM states are initialized through the posterior. The critic is trained on a mixture of real and imagined trajectories. Both the actor and critic networks are optimized via gradient descent. The critic is trained to predict the $\lambda$-return $G_t^\lambda$ from state $z_t, h_t$ by maximizing the log-likelihood under its predicted distribution $p_\psi$:
\begin{equation} \label{eq:critic_loss}
    \mathcal{L}(\psi) = -\sum_{t} \ln p_\psi(G_t^\lambda \mid z_t, h_t).
\end{equation}

The actor loss consists of three terms: a policy-gradient term with a normalized advantage, an entropy bonus, and a smoothness regularizer:
\begin{equation} \label{eq:actor_loss}
\begin{aligned}
    \mathcal{L}(\theta) = -\sum_{t} \bigg(&\!\Big(\big(G_t^\lambda - v_\psi\big) / \max(1,\, S)\Big) \log \pi_\theta \\
    &+ \eta\, \mathrm{H}\!\left[\pi_\theta\right] - \lambda_\text{smooth}\,(\mu_t - \mu_{t-1})^2 \bigg).
\end{aligned}
\end{equation}
The first term is a standard REINFORCE-style update in which the advantage $G_t^\lambda - v_\psi(z_t, h_t)$ is divided by a return-scale estimate $S$ to stabilize training. The entropy term, weighted by $\eta=3 \cdot 10^{-4}$, encourages exploration. The final term penalizes large changes in the policy mean $\mu$ between consecutive steps, weighted by $\lambda_\text{smooth}=5 \cdot 10^{-2}$. We apply this smoothness loss to the policy mean rather than to the sampled stochastic action, following SkyDreamer \cite{verraest2025skydreamer}, since we observe slow convergence when the smoothness reward is applied to the stochastic action directly.

By exploring different actions from visited states, the agent obtains answers to counterfactual questions, while the learned value function enables long-horizon behavior. Although the rollout length appears to bound the planning horizon, the value function informed by the world model can estimate future returns beyond the rollout, allowing the agent to escape locally optimal traps. Fig. \ref{fig:dreaming} shows the decoded depth images from latent imagination using the simulation prior. The actor is trained simultaneously with the RSSM, but their gradients do not update each other. Both the actor and the RSSM with posterior are used for inference and can thus be viewed as a policy collectively.

\subsection{Reward Function}
A key design choice in AirDreamer is the use of primarily sparse rewards. Feedback is provided mainly upon reaching the goal or collision, while the guidance shaping terms are dramatically reduced. This enables emergent yaw control behaviors, avoids reward hacking, removes reliance on human-imposed navigation preferences, and allows the drone to discover adaptive perceptual strategies. Such designs are typically infeasible for prior learning-based methods. Also, the world model enables the agent to propagate such sparse signals through imagined rollouts, making sparse-reward training feasible. The reward $r_t$ used for training consists of two sparse terminal rewards and three small auxiliary shaping terms, defined in discrete time as follows:
\begin{equation}
\begin{aligned}
    r_t =& 10\, r_\mathrm{prog} + 5\, r_\mathrm{safety} + 2\, r_\mathrm{height} \\
         & + r_\mathrm{goal} + r_\mathrm{collision}
\end{aligned}
\end{equation}
where the sparse terminal rewards dominate the supervision signal. $r_\mathrm{goal} = 100$ is given when the agent reaches the goal, terminating the episode. $r_\mathrm{collision} = -5$ is a one-time penalty for a lethal collision, defined as an obstacle within $0.3\,\mathrm{m}$ of the drone's center, which also terminates the episode. Three small auxiliary terms provide minimal training stabilization. $r_\mathrm{prog} = d^{\,t-1}_{xy} - d^{\,t}_{xy}$ is a small progress reward proportional to per-step horizontal distance reduction. $r_\mathrm{safety} = -\max(0,\, 1 - d_\mathrm{obstacle})^2$ is a quadratic penalty active only when an obstacle is within $1\,\mathrm{m}$ of the drone, and zero otherwise; the quadratic form keeps the penalty bounded when the drone must traverse narrow gaps. $r_\mathrm{height} = -(z_\mathrm{goal} - z_\mathrm{drone})^2$ encourages the drone to maintain the goal altitude. The episode is also terminated if the drone exceeds a fixed altitude range, providing a hard vertical constraint.

The large magnitude of $r_\mathrm{goal}$ relative to per-step shaping is intentional. Pure progress shaping induces a local optimum where the agent refuses to detour around obstacles because routing around them temporarily increases $d_{xy}$. The sparse goal reward dominates the learned value function and propagates backward through imagined rollouts, pushing the policy to complete the task even when doing so requires sacrificing short-term progress.

We deliberately give no reward for the camera direction or the yaw angle $\Psi$. Hand-crafting a heading objective would bias the policy toward a specific perceptual strategy that may conflict with the actual task requirements. For instance, a drone backing out of a dead end may need to look opposite to its motion or scan the area, while one entering a corridor may need to look straight ahead. By leaving orientation unconstrained, we allow directional attention to emerge naturally from the visual demands of the task.

\subsection{Sim-to-Real}
We aim to design a method that enables deployment on real drones without significant per-platform parameter tuning. However, the velocity controller and the drone model in the simulator differ from the real-world platforms, creating a sim-to-real gap. A policy trained on a single fixed simulator may fail to transfer when the real drone exhibits different responsiveness or actuation characteristics. 

To address this, we apply domain randomization listed in Table~\ref{tab:domain-rand} to the drone dynamics parameters that govern responsiveness and actuation. The specific quantities are:
\begin{itemize} 
    \item Mass and inertia, which determine the translational and rotational acceleration the drone produces in response to a given thrust and torque. 
    \item Thrust-to-weight ratio, which scales the maximum thrust each rotor can deliver relative to the drone's weight. 
    \item Force-to-moment ratio, which relates each rotor's thrust to the yaw moment it produces; a larger F2M means less yaw authority per unit thrust. \item Drag coefficient $c_d$, which determines the linear body-drag force $F_\mathrm{drag}\propto-c_d\, v$ opposing translational motion. 
    \item Motor spin-up ($\tau_\mathrm{up}$) and spin-down ($\tau_\mathrm{down}$) gains, which control how quickly each rotor responds to commanded throttle changes.
\end{itemize}
The rotor throttle is modeled as a discrete first-order filter, $T_{k+1} = T_k + \tau\,(T^{\,\mathrm{cmd}}_k - T_k)$, where $\tau = \tau_\mathrm{up}$ when the commanded throttle $T^{\,\mathrm{cmd}}$ rises and $\tau = \tau_\mathrm{down}$ when it falls. A larger $\tau$ yields faster rotor response. At the start of each training episode, every randomized parameter is independently resampled. For most parameters, a scaling factor $k \sim \mathcal{U}(\min, \max)$ is drawn from a uniform distribution and multiplied by the simulator's nominal value. Motor gains are sampled from a uniform distribution as absolute values rather than as scaling factors. For parameters with multiple components, each component is sampled independently. For example, the three diagonal entries of the inertia tensor are perturbed separately, and the spin-up and spin-down gains are sampled independently for each of the four rotors. For the depth camera observation, the simulator produces invalid values for pixels outside of the $[0.6,\, 6.0]\, \mathrm{m}$ range. We replace these pixels with the maximum range value (6 m) before passing the image to the policy. We do not simulate any noise or bias in the policy's observation.

\begin{figure}[t]
    \centering
    \includegraphics[width=0.92\linewidth]{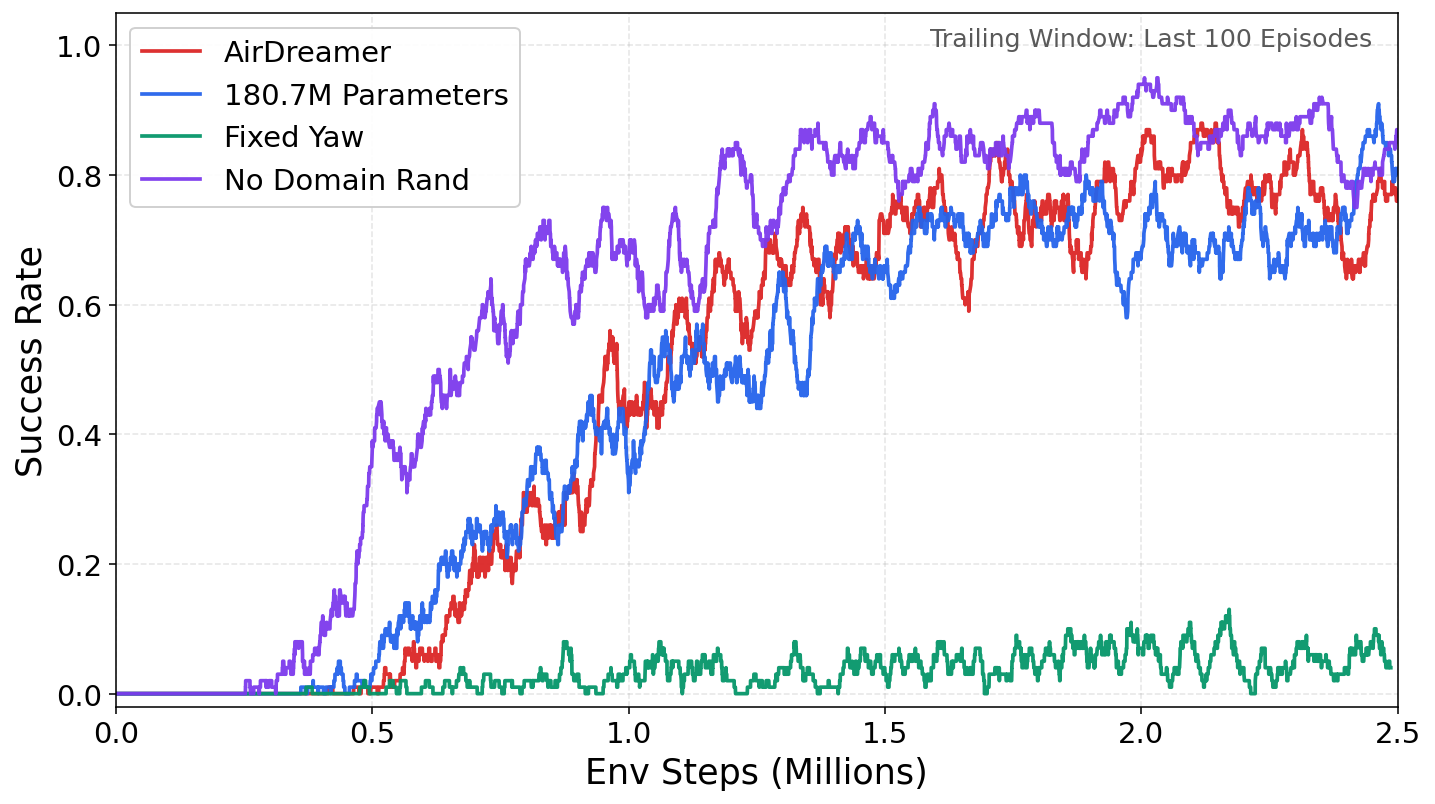}
    \caption{Success rate versus environment steps during training. The label names indicate the parameters varied. For example, in Fixed Yaw, only the yaw angle is fixed, and the run still uses 101.7 M model and domain randomization.}
    \label{fig:train_curve}
\end{figure}

\begin{table}[t]
\centering
\caption{Domain randomization parameters for training}
\label{tab:domain-rand}
\begin{tabular}{@{}lll@{}}
\toprule
Parameter & Min & Max \\
\midrule
Mass scale                    & 0.90 & 1.10 \\
Inertia scale                 & 0.85 & 1.15 \\
Thrust-to-weight scale        & 0.85 & 1.15 \\
Force-to-moment scale         & 0.85 & 1.15 \\
Drag coefficient scale        & 0.75 & 1.25 \\
Motor spin-up gain            & 0.37 & 0.49 \\
Motor spin-down gain          & 0.37 & 0.49 \\
\bottomrule
\end{tabular}
\end{table}

\section{EXPERIMENTS}
\label{sec:exp}
We conduct extensive experiments in both simulation and the real world to evaluate the proposed AirDreamer framework. Specifically, the experiments are designed to answer the following questions: 1. Does AirDreamer outperform state-of-the-art learning-based and classical navigation methods in cluttered and previously unseen environments? 2. Does the sparse reward design enable the agent to escape local optima and exhibit emergent navigation behaviors? 3. Does the trained policy transfer to real drones without per-platform parameter tuning? 

\subsection{Training Details}
\subsubsection{Environment}
A simulation environment shown in Fig. \ref{fig:env} is made to train AirDreamer. We choose OmniDrones \cite{xu2024omnidrones}, a drone simulator based on Isaac Sim 4.1.0 for reinforcement learning. We use the Hummingbird drone model and the LeePositionController (Lee) as the low-level controller. Our environment script takes inspiration from NavRL \cite{xu2025navrllearningsafeflight}. The policy is executed at 20 Hz and the physical simulation runs at 200 Hz.

We do not randomize the positions of static obstacles during training and find doing so does not lead to over-fitting in evaluation cases. We randomly vary the spawn and goal locations and heights, obstacle dimensions, and initial yaw directions.

\subsubsection{Training}
We develop our method based on the original Dreamer V3 codebase in JAX.\footnote{Previous experiments with third-party PyTorch reimplementations produced noticeably worse results.} The JAX model we use has 101.7 million parameters. For deployment, we convert the JAX model to an ONNX model. The inference model for deployment has 57.9 million parameters. The inference model outputs the mean of the policy, justifying the use of smoothness loss.

We train the model for 2.5 million environment steps in total, equivalent to 34.7 hours of experience, and use the model at 2.25 million environment steps. The training curve (red) is shown in Fig. \ref{fig:train_curve}. Before 0.5 million env. steps, it learns to adapt to different drone parameters and to perform maneuvers. We limit the replay buffer size to 1 million steps to prevent RAM overflow. The training and simulation take place on a server with 128GB RAM and two NVIDIA RTX 4090 D (24GB). The complete training run takes 38 hours and 43 minutes. 

\begin{table}[t]
\centering
\caption{Success rate benchmark}
\label{tab:bench_stats}
\begin{tabular}{@{}lcccc@{}}
\toprule
Method & Success Rate (\%) & Avg Speed (m/s)\\
\midrule
AirDreamer & \textbf{59.3 $\pm$ 15.8}  & 0.45\\
DepthNav   & 54.0 $\pm$ 13.9 & 0.48\\
NavRL      & 9.3 $\pm$ 3.9 & 0.31\\
EgoPlanner & 6.7  $\pm$ 6.3  & 0.57\\
\bottomrule
\end{tabular}
\end{table}

\subsection{Comparison to State-of-The-Art Methods}
AirDreamer is benchmarked against DepthNav \cite{lee2025quadrotor}, NavRL \cite{xu2025navrllearningsafeflight}, and EgoPlanner \cite{zhou2020ego}. While MAVRL \cite{yu2024mavrllearnflycluttered} is also targeted at end to end navigation for drones, it uses a low-level controller without open access, and our attempt to use a substitute did not preserve the original method's performance. We therefore exclude it to avoid an unfair comparison. Table \ref{tab:bench_stats} summarizes the success rates and the average speeds of successful runs for each method. Each success rate is obtained from a total of 150 runs across 5 maps with equal density in the XTDrone simulator \cite{xiao2020xtdrone}. Real-world experiments showed large, method-dependent swings in success rate, because flight-control fine-tuning is required for deployment and is not fully open-sourced across baselines. To remove this confound, we report success rates in XTDrone and real-world experiments are used for qualitative analysis.

The results show that AirDreamer performs 5.3\% better success rate than the best baseline and 52.6\% higher success rate than classical optimization-based methods. We observe a trend that all methods achieve similar success rates above 90\% in simple environments, while AirDreamer shows a larger performance advantage as environment complexity increases. The classical method \cite{zhou2020ego} shows a low success rate due to the local observation constraint. Such methods can only guide by the goal direction and easily falls into local optima planning. NavRL imposes a constraint on yaw control, which prevents it from finding paths that are not directly visible, resulting in a low success rate. DepthNav trains the policy reinforcement learning, which suffers from the OOD problem and shows a lower success rate. Also, it needs dense reward to guide the training, which introduces human-preferred paths and also leads to local optima when tested in randomly generated environments. 

\begin{figure}[t!]
    \centering
    \includegraphics[width=0.85\linewidth]{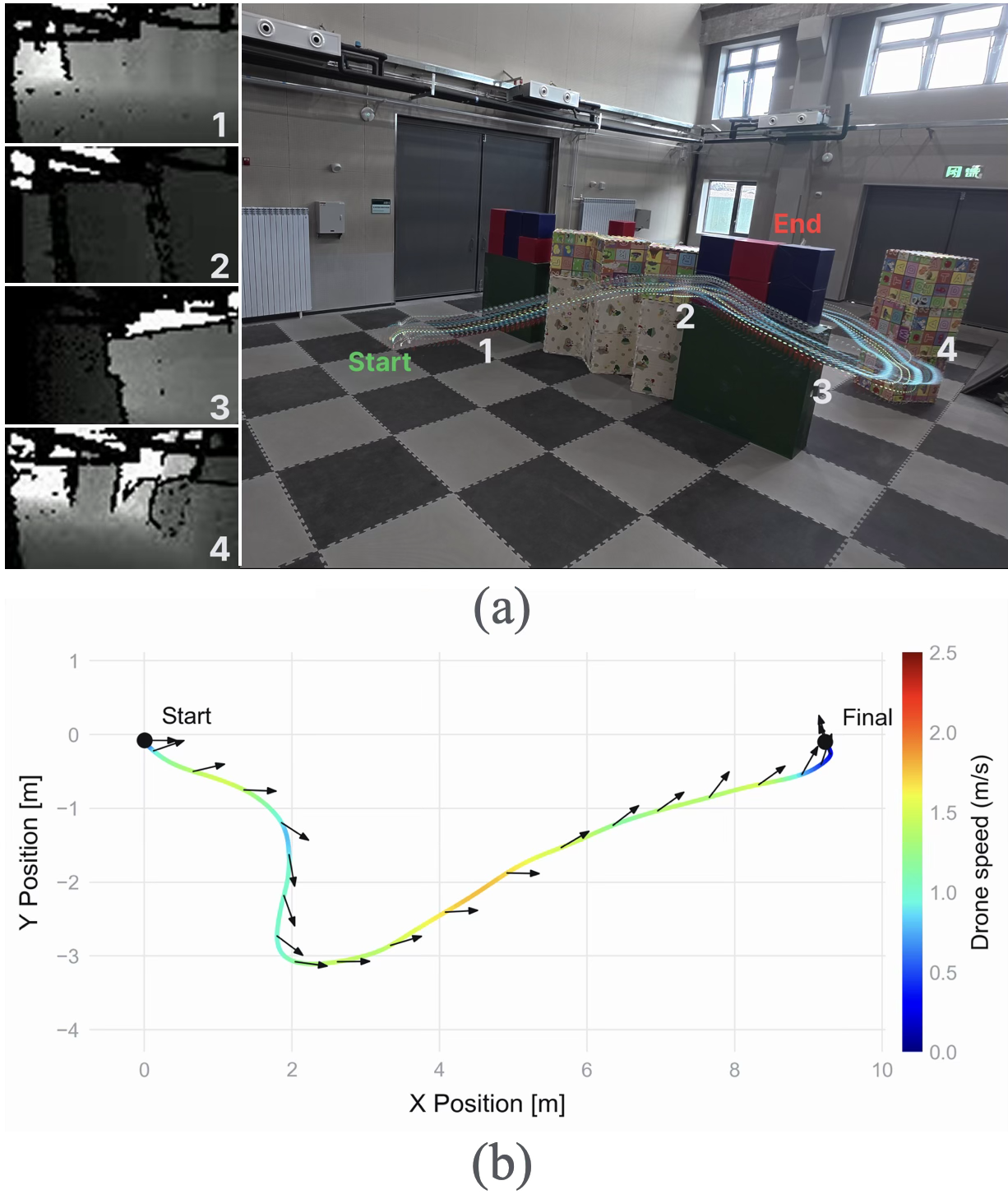}
    \caption{(a) Trajectory visualization by stacking pictures from the run in 5m Wall layout. The depth images at a few locations are shown. (b) Trajectory visualization by drawing the recorded XY position of the drone in 5m Wall layout. The arrows indicate the camera direction.}
    \label{fig:wall_traj}
\end{figure}

\subsection{Real-world Performance}
To evaluate the robustness and sim-to-real transferability of AirDreamer, we deploy the system on a real quadrotor with a 280 mm motor-to-motor span. State estimation is provided by FAST-LIO2 \cite{xu2021fastlio2}, depth observations are captured by a RealSense D455, the low-level flight control runs on a Pixhawk 6X Pro with PX4 v1.14.3, and our method executes on a Jetson Orin NX Super (16GB). The lidar installed on the drone is only used for localization and is not used by the navigation policy. Our 57.9M-parameter policy network reaches up to 88 Hz inference on the Jetson and is deployed at 20 Hz on a ROS node. The task is completed when the drone arrives within 1 m of the goal. As shown in Fig. \ref{fig:cover}, six demonstration runs are conducted across diverse scenarios, including cluttered obstacle avoidance (Forest), precise flight through a narrow gap (1m Passage), navigation around a large obstacle (5m Wall), escaping from a local optimum (U-shape), and generalization to an unseen obstacle type (C-shape). The drone successfully completes all scenarios, and flight statistics averaged over three trials per scenario are summarized in Table \ref{tab:flight-stats}.

\begin{figure}[t!]
    \centering
    \includegraphics[width=0.85\linewidth]{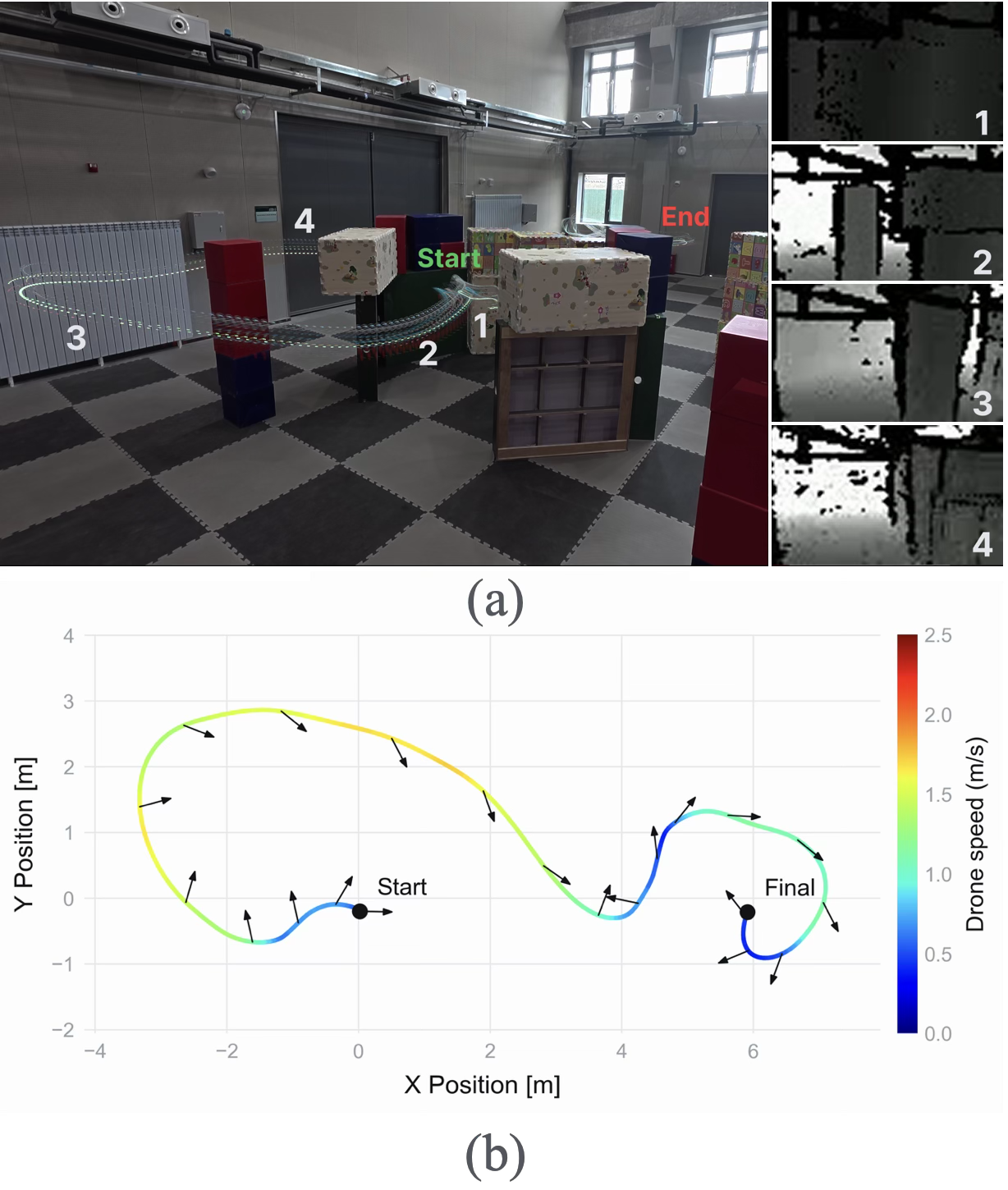}
    \caption{(a) Trajectory visualization by stacking pictures from the run in C-shape layout. The depth images at a few locations are shown. (b) Trajectory visualization by drawing the recorded XY position of the drone in C-shape layout. The arrows indicate the camera direction.}
    \label{fig:c_traj}
\end{figure}

\begin{table}[t]
\centering
\caption{Per-run flight statistics}
\label{tab:flight-stats}
\begin{tabular}{@{}lcccc@{}}
\toprule
\raisebox{1.2ex}{Run} & \shortstack[c]{Avg spd\\(m/s)} & \shortstack[c]{Max spd\\(m/s)} & \shortstack[c]{Direct dist\\(m)} & \shortstack[c]{Path dist\\(m)}\\
\midrule
5m Wall   & 1.1 & 1.8 & 9.0  & 12.6 \\
1m Passage       & 1.1 & 1.6 & 9.0  & 12.2 \\
U-shape    & 1.0 & 1.4 & 8.0  & 11.5 \\
C-shape    & 1.0 & 1.8 & 6.0  & 21.3 \\
Forest-forward   & 1.1 & 1.7 & 10.0 & 12.1 \\
Forest-backward   & 1.1 & 1.7 & 10.0 & 12.0 \\
\bottomrule
\end{tabular}
\end{table}

During the runs, the drone can perform agile sideway maneuvers when an obstacle is in the way. Here, we analyze two representative runs to showcase the capability of AirDreamer. The first run is long Wall, as shown in Fig. \ref{fig:wall_traj} (a). A 5 meter wall is placed between the drone and the goal, with both edges initially outside the drone camera's field of view. Without an immediately visible path forward, the policy must yaw the drone to scan the environment and identify a viable detour. As shown in Fig.~\ref{fig:wall_traj} (b), the drone first yaws left to scan the environment. Because the left side offers insufficient clearance, the policy then steers the drone to the right. Upon approaching the wall, the drone slows down by approximately 0.5 m/s and exhibits wall-following behavior. Behind the wall lies a cuboid obstacle, which the policy actively yaws toward in flight. This demonstrates emergent yaw control learned from sparse rewards without explicit directional guidance. Despite the invalid pixels and the vision blockage caused by the chassis, the policy adapts robustly to this complex unseen scenario.

The second run is a C-shaped obstacle, as shown in Fig. \ref{fig:c_traj} (a). During training, the drone has only seen U-shaped obstacles. So this test is out-of-distribution for the policy. Also, the drone is initially placed in the obstacle's cavity and has to rotate its vision to find the exit. Despite this, the drone successfully escapes from the trap by first flying away from the goal for 3 meters, ignoring the local optimum. Notably, the policy first controls the drone's yaw to expand its visual coverage, revealing a previously unseen passage, as shown in Fig. \ref{fig:c_traj} (b). The policy then control the done move laterally with the camera oriented toward the obstacles. After escaping from the trap, the policy command speeds up to 1.7 m/s from 0.5 m/s while maintaining the obstacles in view. Previous methods that depend on explicit directional rewards or pre-defined waypoints cannot navigate this scenario.

\subsection{Ablation}
To further demonstrate the contribution of each component of AirDreamer, we conduct an ablation study on three factors, model capacity, domain randomization, and yaw control. 

The training curves of our nominal model and the three variants are shown in Fig.~\ref{fig:train_curve}. Scaling the policy network from 101.7 M to 180.7 M parameters yields no measurable improvement in success rate, indicating that AirDreamer's performance comes from the framework design rather than raw model capacity. Disabling domain randomization improves the success rate by about 10\%, but the resulting policy cannot be deployed on real drones. This trade-off between in-simulation performance and sim-to-real transferability is consistent with prior observations \cite{ferede2025onenet}. Finally, fixing the yaw angle substantially degrades performance, confirming that allowing the policy to control yaw is essential for the agent to align its perception with task demands.

\section{CONCLUSION}
AirDreamer is a concise, end-to-end framework for drone navigation in unknown, cluttered environments, requiring neither global maps nor pre-defined rules. Extensive simulation and real-world experiments demonstrate state-of-the-art performance, with AirDreamer achieving a  5.3\% improvement in success rate over the strongest baseline and successfully navigating obstacle configurations never seen during training, where previous methods fail. The emergent behaviors, such as active sideways scanning to compensate for the limited camera field of view and wall-following when no opening is visible, suggest that the policy has internalized a meaningful understanding of depth observations rather than memorizing training trajectories. We believe this framework can be extended to dynamic obstacles, multi-robot systems, and target tracking without significant effort. The framework's end-to-end neural architecture and minimal sensor requirement make it well-suited for resource-constrained platforms and future edge-computing hardware, with potential applications in industrial inspection, search and rescue, and autonomous delivery where pre-mapping is impractical.

\bibliographystyle{IEEEtran}
\bibliography{ourbib}

\end{document}